\documentclass[journal]{IEEEtran}

\usepackage{color}
\usepackage{graphicx}
\usepackage{amssymb}
\usepackage{epstopdf}
\usepackage{amsmath}
\usepackage{bm}
\usepackage{amsmath}
\usepackage[lined,boxed,linesnumbered,noend]{algorithm2e}
\usepackage{multirow}
\usepackage{cite}

\usepackage{array}

\usepackage{bigstrut}
\newcolumntype{C}[1]{>{\centering\let\newline\\\arraybackslash\hspace{0pt}}m{#1}}

\title{\LARGE \bf Efficiently Improving and Quantifying Robot Accuracy In Situ}

\author{Karl~Van~Wyk,
        Joe~Falco,
        Geraldine~Cheok
\thanks{}
\thanks{\newline
Karl Van Wyk, Joe Falco, and Geraldine Cheok are with the National Institute of Standards and Technology, Gaithersburg, MD, USA,  (e-mail: karl.vanwyk@nist.gov; joseph.falco@nist.gov; cheok@nist.gov). \newline
Official contribution of the National Institute of Standards and Technology; not subject to copyright in the United States.}
}

\begin{document}

\maketitle
\thispagestyle{empty}
\pagestyle{empty}

\begin{abstract}
The advancement of simulation-assisted robot programming, automation of high-tolerance assembly operations, and improvement of real-world performance engender a need for positionally accurate robots. Despite tight machining tolerances, good mechanical design, and careful assembly, robotic arms typically exhibit average Cartesian positioning errors of several millimeters. Fortunately, the vast majority of this error can be removed in software by proper calibration of the so-called ``zero-offsets'' of a robot's joints. This research developed an automated, inexpensive, highly portable, \textit{in situ} calibration method that fine tunes these kinematic parameters, thereby, improving a robot's average positioning accuracy four-fold throughout its workspace. In particular, a prospective low-cost motion capture system and a benchmark laser tracker were used as reference sensors for robot calibration. Bayesian inference produced optimized zero-offset parameters alongside their uncertainty for data from both reference sensors. Relative and absolute accuracy metrics were proposed and applied for quantifying robot positioning accuracy. Uncertainty analysis of a validated, probabilistic robot model quantified the absolute positioning accuracy throughout its entire workspace.  Altogether, three measures of accuracy conclusively revealed multi-fold improvement in the positioning accuracy of the robotic arm. Bayesian inference on motion capture data yielded zero-offsets and accuracy calculations comparable to those derived from laser tracker data, ultimately proving this method's viability towards robot calibration. \newline



\end{abstract}

\section{Introduction}
\label{sec:introduction}
\IEEEPARstart{T}{wenty} years ago, researchers concluded that metrology systems for robot calibration were not economical. In fact, it was stated that ``...the development of a system that could combine these characteristics, but at a low-cost, would fill an important void in the automation industry \cite{Hidalgo1998}." Despite advancement in these metrology systems, many remain cost-prohibitive at a time of increasing robotic automation adoption rates \cite{IFR}. With a growing need for rapid robot deployment, application-level robot programming strategies are relying more heavily on offline or simulation-assisted environments that require accurate modeling of the robots themselves and their workcell environments. These programs can then be rapidly generated with improved collision-avoidance and fine-assembly behaviors, and can be identical for all robots of the same make and model. This programming paradigm is in stark contrast with lead-through programming, where robot positions are manually configured, logged, and replayed on a per-robot basis \cite{Pan2012}. 

Furthermore, many compensatory behaviors for robots, e.g., force control or planning, already exist that seek to mitigate the effects of positional inaccuracies. However, despite their best efforts, performance degradation, in terms of likelihood of success or completion time, still exists and scales with positioning error. Van Wyk et al. \cite{vanwyk2018comppeg} applied force control methods, including that of a complex robotic hand, to successfully perform a peg-in-hole task in the presence of hole position error. Results indicated that all control strategies for completing the peg-in-hole task experienced statistically significant escalations in completion times of the task with increased hole position error. Kaipa et al. \cite{kaipa2016addressing} combined simulation and probabilistic methods to generate singulation plans of binned parts. Data revealed that as the positional error of the part increased, the probability of a successful singulation decreased. Mahler et al. \cite{mahler2016dex} also reported that grasp planning quality on objects was inversely correlated with magnitude of object pose error despite sophisticated algorithms that compensated for these errors. Liu and Carpin  \cite{liu2016kinematic} showed how time-to-convergence for grasp planning positively correlated with increased levels of uncertainty, and likelihood of force closure negatively correlated with increased levels of uncertainty. Conclusively, the performance of both open- and closed-loop methods for programming and controlling robots for task-level applications can benefit from intrinsically more accurate robots.

Various factors contribute to the positional accuracy of robots, including machining tolerances, motor and gearhead precision, encoder resolution, structural loading (gravitational, inertial, thermal), position controllers, and joint zero-offset calibration. Of these factors, research has indicated that more than 90\,\% of positional inaccuracy is caused by a robot's zero-offset parameters \cite{Judd1990,Conrad2000}. These zero-offsets are constants added to the joint positions that make the manufactured robot's kinematics more accurately reflect its theoretical kinematics. The Cartesian accuracy of a robot arm is highly sensitive to small errors in joint-offsets (typically angular), with nonlinearly propagated effects to a robot's tool center point (TCP). This effect is further exacerbated by robots with greater reach. Fortunately, of the previously listed factors, the zero-offset parameters are the most readily modifiable. In fact, customers are often encouraged to update their robots' zero-offsets, henceforth referred to as "remastering", by robot manufacturers after robot delivery and periodically due to drift and wear-and-tear \cite{Gao2014}.

\begin{figure}
\centering
\includegraphics[width=1\linewidth]{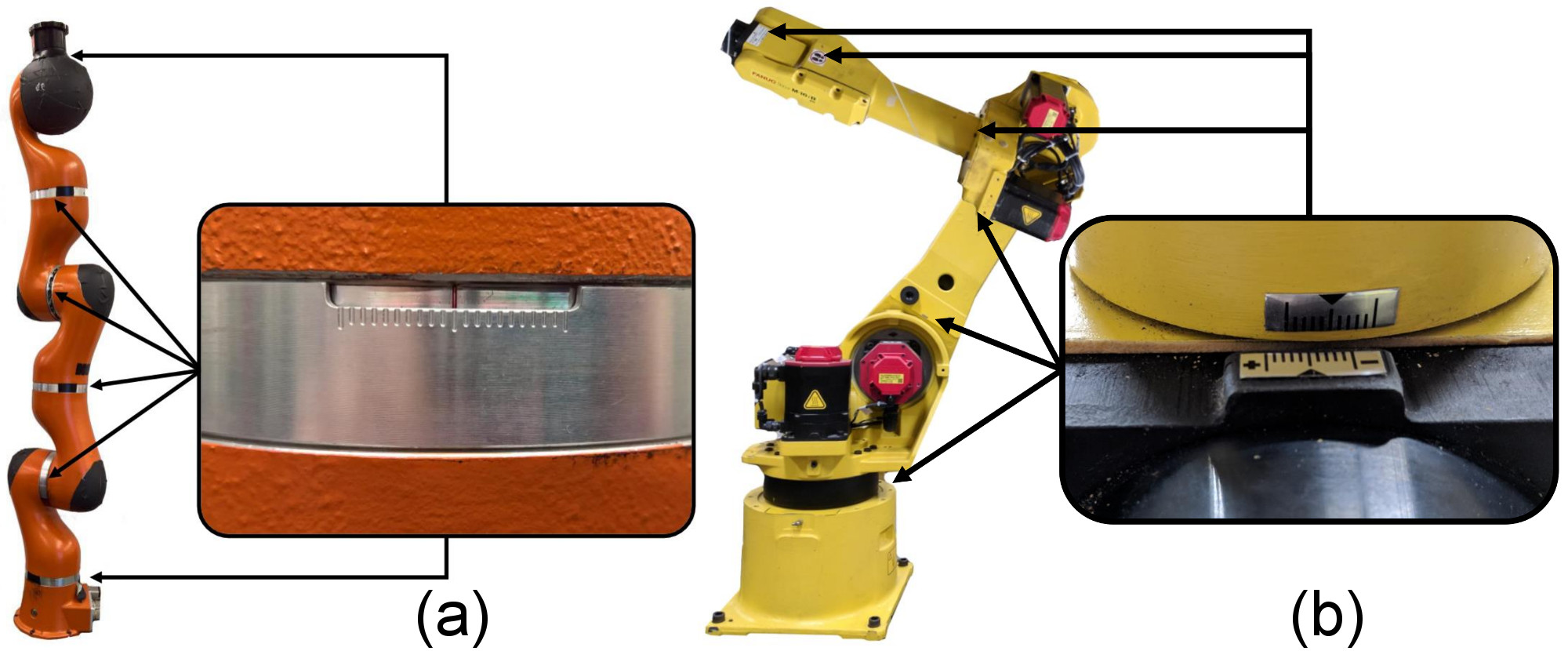}
\caption{Vernier scales on (a) KUKA and (b) Fanuc robot arms.}
\label{fig:ticks}
\end{figure}

Possessing an economical and effective method for remastering robots is paramount to improving and maintaining their overall positioning accuracy. Counter-intuitively, some robot manufacturers do not provide methods for users to remaster their robots on-site. Moreover, some existing, rudimentary approaches to remastering robots involve visual alignment of Vernier scales at the robot joints (see Fig. \ref{fig:ticks}). Albeit simple, this process is error-prone due to visual inspection, assembly tolerances, and potential warping or shifting of the guide band over time. As revealed herein, remastering the KUKA robot with its Vernier scales yielded average Cartesian positioning errors of several millimeters with extreme cases as large as 10 mm.

There exist many methods in the literature for remastering robots, all of which require external sensors. For instance, many modern approaches leverage laser trackers that are notoriously expensive, bulky, and require operator training, expertise, and periodic recalibration. Mustafa et al. \cite{Mustafa2015} used an iterative parameter identification approach using the product-of-exponential (POE) formula and data collected by a laser tracker. The average robot positioning error was reduced to 0.29 mm from 5.71 mm. Tao et al. \cite{Tao2015} extended this approach by applying both POE modeling (for kinematic errors) and a neural network (for non-kinematic errors), which reduced positioning errors from over 1 mm to 0.34 mm. Choi et al. \cite{Choi2016} applied POE with data collected from a laser tracker as well to reduce position errors from 1 cm - 3 cm to within 1 cm. Jiang et al. \cite{jiang2017new} applied, in-sequence, an Extended Kalman Filter and particle algorithm to optimize all Denavit-Hartenberg (D-H) parameters of the robot from data collected by a laser tracker. The average robot positioning error was reduced to 0.26 mm from 3.14 mm.

Fortunately, some methods exploit lower-cost sensory systems as well, such as lasers and position sensitive detectors (PSD), or theodolites. Chen et al. manually guided a robot along a laser line that centers on a pin hole at the robot end-effector and calculated zero-offset parameters such that the end-effector distances to the line are minimized \cite{Chen2008}. Robot positioning errors were reduced to 0.572 mm from over 3 cm. In several works, Y. Liu et al. leveraged a low-cost PSD and laser pointer to automate the process of calculating zero-offset parameters through numerical optimization \cite{Liu2009_1,Liu2009_2,Gao2014,Liu2014}. Some results showed reduction in positioning errors from over 3 cm to within 0.898 mm. 

Clearly, there exist many different algorithmic approaches and sensors for remastering a robot for significantly improved Cartesian accuracy. However, attractive features of future-forward remastering methods should prioritize increased efficiency in terms of cost, portability, and convenience, while maintaining competitive accuracy gains. These priorities are particularly necessary for reducing the barrier to entry for small- and medium-sized enterprises \cite{marvel2015nistir}. Accordingly, this research provides an efficient remastering solution with these features alongside three proposed metrics for evaluating robot accuracy. The contributions of this research are organized to convey these advancements as follows. Section \ref{sec:hardware} identifies key hardware components for conducting experiments. Section \ref{sec:data_collection} discusses the data collection process. Section \ref{sec:optimization} mathematically formulates the joint posterior of crucial parameters, and calculates the optimized zero-offsets and their uncertainty with Bayesian optimization. Section \ref{sec:quantifying_accuracy} proposes three metrics for quantifying robot accuracy, including a novel, theoretical calculation based on model uncertainty. Section \ref{sec:validation} validates experimental findings with accuracy measurements on a validation dataset. Finally, Section \ref{sec:discussion} frames the primary research results, their ramifications, and suggests future improvements.

\section{Hardware and Setup}
\label{sec:hardware}

The designed setup for remastering a robot was minimally invasive without the typical requirement of high-tolerance, marker-mounted plates at the base of the robot for registering the reference measurement system with the robot \cite{jiang2017new}. Therefore, robot remastering can occur \textit{in situ}, with only a single reference system mounted in proximity to the robot as shown in Fig. \ref{fig:setup}. For these experiments, two reference measurement systems were used for benchmarking and validation purposes, but this is not required in practice. Coordinate systems for the various reference frames are defined as follows: 1) $R$ for robot base, 2) $T$ for robot tool, 3) $M$ for motion capture system, 4) $L$ for laser tracker, and 5) $N$ for general reference system. 

\subsection{Robot}
\begin{figure}
\centering
\includegraphics[width=1\linewidth]{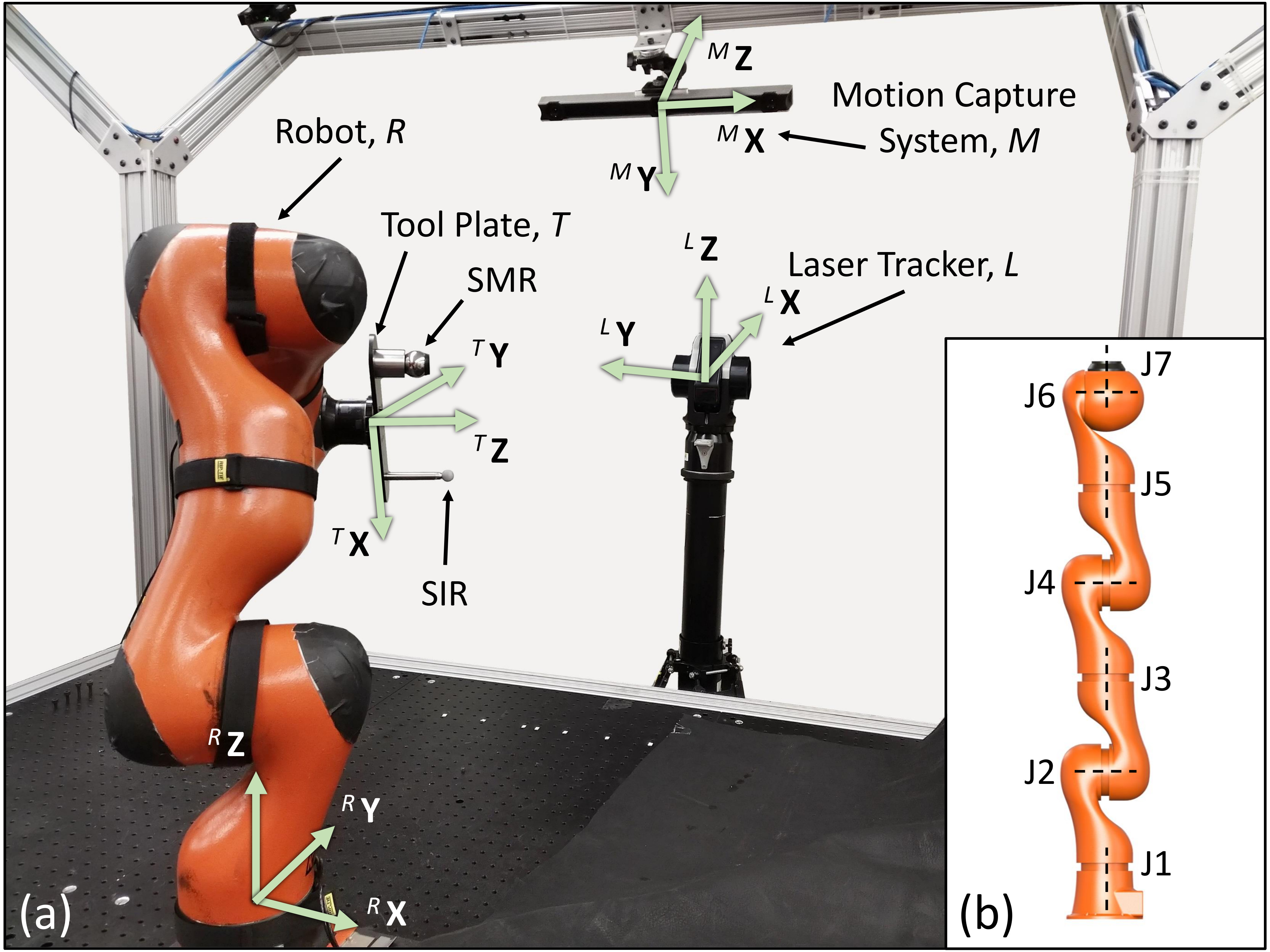}
\caption{Hardware setup and placement of (a) coordinate systems for robot, tool plate, motion capture system, and laser tracker, and (b) robot axes J1 through J7.}
\label{fig:setup}
\end{figure}

The robotic system under test was a 7 degree-of-freedom (DoF), KUKA arm that operated in high-stiffness, joint position control to maximize free-space positioning accuracy. A custom tool plate that housed two types of markers was attached to the end of the arm via a commercial quick-change connector (see Fig. \ref{fig:setup}). This connector allowed for a fast connection to any robot arm with the reciprocal connector. A spherically-mounted retroreflector (SMR) was attached to one end of the tool plate at a 100 mm offset from the plate center point (plate center and J7 axis intersect). A spherical infrared reflector (SIR) was attached to the opposite end of the tool plate at a 100 mm offset as well. The SMR and SIR allowed for 3D position measurement with a laser tracker and motion capture system, respectively. Offsetting the markers from J7 ensured that marker positioning was sensitive to the angular position about J7. Mounting both types of markers allowed for measurement, zero-offset optimization, and analysis with both the unconventional motion capture system and the benchmark laser tracker. Before experimentation, the robot was manually remastered by visually aligning the Vernier scales and calibration marks (see Fig. \ref{fig:ticks}) across all seven joints as described by the robot manufacturer.

\subsection{3D Reference Sensors}
The OptiTrack Trio is a low-cost, pre-calibrated, plug-and-play motion capture system for measuring the 3D locations of SIRs. With a resolution of 640 pixels $\times$ 480 pixels for each of its three cameras, the device possessed sufficient spatial resolution as a reference measurement system. Preliminary investigation revealed that the measurement noise produced from this device had a standard deviation within 0.03 mm per axis. The accuracy of the device is unknown; however, the more significant requirement is that the device's perception of distance (dependent on sensor hardware and calibration fidelity) does not significantly distort as a function of marker location. As later revealed in Section \ref{sec:optimization}, any Cartesian calibration inaccuracy in this device was compensated for by simultaneously optimizing the registration mapping between the sensor and base location of the robot alongside the zero-offset parameters. The device can be mounted anywhere, but a slightly overhead angle can reduce the likelihood of visual occlusion. In these experiments, measurement of SIR positions were seamless and automatic - free of issues concerning occlusion or temporary visual marker loss. Data collection was expedited by the sensor's NatNet User Datagram Protocol that allowed for streaming data to a remote client.

The API T3 laser tracker had a measurement range of 80 m with a resolution of 0.1 $\mu$m and accuracy $\pm$ 15 $\mu$m in its Absolute Distance Meter (ADM) measurement mode. The ADM mode was used in these experiments to allow measurement after semi-automatic re-acquisition of a lost marker. Beam breaking (marker loss) was inevitable during large joint reconfiguration changes of the robot arm as discussed in Section \ref{sec:data_collection}. Although the ADM mode ultimately provided a solution path for position measurement for this application, the partly manual re-acquisition of markers impeded the rate of data collection. Finally, the laser tracker can be placed arbitrarily so long as it has line-of-sight to the SMR. In this case, the laser tracker was placed below the motion capture system to ensure a similar viewpoint. The significance of using a laser tracker in these experiments included benchmarking and validation of motion capture measurements. Specifically, zero-offset optimization was performed and compared between data acquired from the laser tracker and the motion capture system. Agreement in joint-offset solutions and accuracy calculations between the laser tracker and motion capture system confirmed the latter device as sufficient for robot remastering.

\section{Data Collection}
\label{sec:data_collection}
The first step in the remastering process required collecting Cartesian positioning data as measured by both the robot and an extrinsic reference measurement system. The optimization process will then adjust the robot's zero-offsets to maximize the agreement in Cartesian positioning as reported by the robot and a reference device. To increase the value of data acquired, the robot's end-effector was systematically repositioned throughout its workspace while measuring its Cartesian positions. The following discusses the exact process taken for data collection and along with important considerations.

\subsection{Robot Positioning}
To streamline the data collection process and obtain high-quality measurements, visual occlusion (full or partial) of both the SMR and SIR must be avoided. This requirement was met by ensuring that the Z-axis of the robot end-effector, ${}^T \mathbf{Z}$, pointed towards the origin of the laser tracker for any Cartesian pose commanded to the robot. This heading was biased towards the laser tracker as it can more easily lose sight of the SMR. In contrast, measurement by the motion capture system is more forgiving since the SIR allows for larger view angles. The mounted proximity of the laser tracker and motion capture system ensured unobstructed views of both the SMR and SIR. 

To begin, a single Cartesian pose expressed in $R$, ${}^R\mathbf{r}_d = [{}^R x_d, {}^R y_d, {}^R z_d, \gamma_d, \theta_d, \phi_d]^T \in \mathbb{R}^{6 \times 1}$, was defined by its translations, ${}^R\mathbf{p}_d = [{}^R x_d, {}^R y_d, {}^R z_d]^T \in \mathbb{R}^{3 \times 1}$, and ZYX Euler rotations, $\mathbf{\Theta}_d = [\gamma_d, \theta_d, \phi_d]^T \in \mathbb{R}^{3 \times 1}$. A Cartesian pose can also be expressed as a homogeneous transformation matrix, ${}^R \mathbf{T}_d \in \mathbb{R}^{4 \times 4}$, as

\begin{equation} \label{eq:pose_matrix}
{}^R \mathbf{T}_d = 
\begin{bmatrix}
 & {}^R \mathbf{R}_d & & {}^R\mathbf{p}_d \\
 0 & 0 & 0 & 1
\end{bmatrix},
\end{equation}

\noindent where ${}^R \mathbf{R}_d = [{}^R \mathbf{\hat{x}}_d, {}^R \mathbf{\hat{y}}_d, {}^R \mathbf{\hat{z}}_d] \in \mathbb{R}^{3 \times 3}$ is the rotational matrix form of the ZYX Euler rotations. 

Candidate Cartesian poses were pseudo-randomly generated via the Latin Hypercube sampler (LHS) in four-dimensional space and mathematically converted to a full, six-dimensional Cartesian pose, ${}^R \mathbf{r}_d$. Specifically, LHS produced values for $\{{}^R \theta_z,r,{}^R z_d,{}^T \theta_z\}$, where ${}^R \theta_z$ is an angle about ${}^R \mathbf{Z}$, $r$ is a radius in the ${}^R \mathbf{X}-{}^R \mathbf{Y}$ plane emanating from the origin of $R$, ${}^R z_d$ is previously defined, and ${}^T \theta_z$ is an angle about ${}^T \mathbf{Z}$. The radial magnitudes $r$ ranged from 200 mm to 500 mm, and ${}^R \theta_z$ ranged from 0 to 2$\pi$ rad. The polar coordinates $r$ and ${}^R \theta_z$ were converted to Cartesian coordinates, yielding ${}^R x_d$ and ${}^R y_d$. Values for ${}^R z_d$ were generated from 400 mm to 800 mm. The sampled volume for these translational positions primarily existed within the robot's \textit{dexterous} workspace, i.e., the volume within which the robot's end-effector can be arbitrarily reoriented. Operating within this volume increased the likelihood that multiple physically realizable joint configurations existed for a candidate Cartesian pose. Moreover, the resulting locations of the reference markers existed within full view of both the laser tracker and all three cameras of the motion capture system.

With ${}^T \mathbf{Z}$ pointing towards the origin of the laser tracker, candidate $\lbrace \gamma_d, \theta_d, \phi_d \rbrace$ depended on the relative location of the end-effector with respect to the laser tracker. Therefore, the laser tracker was first registered to the base of the robot by manually positioning the end-effector (with markers) in three different locations. At each location, the SMR position was recorded in both $R$ (calculated with robot forward kinematics) and $L$. The registration process outlined in \cite{Marvel2016} was used to calculate a homogeneous transformation from the laser tracker to the robot, ${}^R_{L} \mathbf{T} = \begin{bmatrix} & {}^R_{L} \mathbf{R} & & {}^R \mathbf{p}_{O,L} \\ 0 & 0 & 0 & 1 \end{bmatrix}$, where ${}^R_{L} \mathbf{R}$ and ${}^R \mathbf{p}_{O,L}$ are the rotation matrix and origin of the laser tracker expressed in $R$. Given ${}^R \mathbf{p}_{O,L}$ and ${}^R \mathbf{p}_d$, ${}^R \mathbf{\hat{z}}_d$ was calculated as

\begin{equation} \label{eq:zaxis}
{}^R \mathbf{\hat{z}}_d = \frac{{}^R\mathbf{p}_{O,L}-{}^R \mathbf{p}_d}{||{}^R\mathbf{p}_{O,L}-{}^R \mathbf{p}_d||},
\end{equation}

\noindent which forced ${}^T \mathbf{Z}$ to point towards the laser tracker origin. Next, ${}^R \mathbf{\hat{x}}_d$ was calculated by

\begin{equation} \label{eq:xaxis}
\begin{array}{ll}
{}^R\mathbf{\hat{x}}_d & = {}^R\mathbf{\hat{z}}_d \times {}^R\mathbf{Z} \\
& = {}^R\mathbf{\hat{z}}_d \times \begin{bmatrix}
0 \\
0 \\
1
\end{bmatrix}
\end{array},
\end{equation}

\noindent where ${}^R\mathbf{\hat{z}}_d$ and ${}^R\mathbf{Z}$ are never parallel in this particular setup. Alternatively, a different vector may also be used instead of ${}^R\mathbf{Z}$ so long as the two vectors are not parallel. Finally, ${}^R\mathbf{\hat{y}}_d$ is calculated as

\begin{equation} \label{eq:yaxis}
{}^R\mathbf{\hat{y}}_d = {}^R\mathbf{\hat{z}}_d \times {}^R\mathbf{\hat{x}}_d
\end{equation}

\noindent to establish a right-handed, orthonormal coordinate system. Unfortunately, this formulation established a spatial correlation between the directionality of ${}^R\mathbf{\hat{x}}_d$ and ${}^R\mathbf{\hat{y}}_d$ and the translational position of the end-effector. To preserve the randomness of the Cartesian poses, ${}^R\mathbf{R}_d$ was updated by an additional rotation of ${}^T \theta_z$ about ${}^R \mathbf{\hat{z}}_d$ (previously defined),

\begin{equation} \label{eq:rotation_new}
{}^R\mathbf{R}_d(\theta_z) = {}^R\mathbf{R}_d 
\begin{bmatrix}
\cos({}^T \theta_z) & -\sin({}^T \theta_z) & 0 \\
\sin({}^T \theta_z) & \cos({}^T \theta_z) & 0 \\
0 & 0 & 1
\end{bmatrix}.
\end{equation}

\noindent Given ${}^R\mathbf{R}_d$, well-known analytical equations were used to calculate the corresponding ZYX Euler angles, $\lbrace \gamma_d, \theta_d, \phi_d \rbrace$ \cite{spong2006robot}.

The process for pseudo-randomly generating ${}^R \mathbf{r}_d$ was conducted twice to yield one set of candidate poses for optimization and another set for validation. All poses were sequentially commanded to the robot and checked for visual occlusion, reachability, and arm collisions. Any problematic poses were discarded, which yielded approximately three dozen poses (from an initial 60) for both optimization and validation, separately. This quantity of data yielded good convergence during optimization. Experimental data were collected with the motion capture system within 20 min per pose set (optimization and validation). Acquisition time depended on the speed setting of the robot (set at 30 \%), settling time for the robot once it reached a desired configuration (set to 3 sec), and commanded order of Cartesian poses (ordered based on proximity to current pose). In contrast, data collection took approximately 3.5 h with the laser tracker to record the optimization dataset due to the semi-manual process of re-acquiring markers after beam breaking. Regardless, data collection times could be improved with faster arm speed settings and smaller delays for robot settling.

\subsection{Analytical Inverse Kinematics}
All prior approaches to optimizing zero-offsets involved re-positioning the robot throughout its workspace along some geometric path (e.g., line or circle), about single points, or randomly. However, an overlooked feature of a robot is its ability to achieve a particular Cartesian pose with potentially many different joint positions. The process of calculating a set of joint positions for a robot that yields a desired Cartesian pose is called inverse kinematics. Moreover, analytical inverse kinematics (AIK) generate exact equations for calculating \textit{all} possible joint position solutions for any given Cartesian pose. Consequently, for every candidate Cartesian pose in the optimization and validation sets, the robot was commanded to all possible joint solutions (using AIK), forcing the robot to undergo large configuration changes and yielding immediate disparities in the resulting Cartesian position. These differences are directly measurable with a reference system and predominantly originate from errors in joint-offsets that propagate differently for different joint solutions. This characteristic immediately rendered a clear picture of the positioning error magnitude as discussed in Section \ref{sec:dist_accuracy_measure}. To obtain this metric, a custom analytical inverse kinematics library was created using existing methods \cite{spong2006robot} to command the robot to all possible joint solutions for every valid Cartesian pose. Overall, 36 Cartesian poses for optimization yielded 173 joint configurations, and 32 Cartesian poses for validation yielded 156 joint configurations. 

\section{Optimization}
\label{sec:optimization}
The zero-offsets for the robotic arm were calculated by sampling from a derived posterior distribution of these parameters with the well-known Metropolis algorithm \cite{hastings1970monte} given experimental data. This algorithm was chosen so that the most-likely estimates (MLE) for the zero-offsets could be obtained along with their uncertainty in the form of probabilistic distributions. The MLE zero-offsets were directly applied to the robot to complete robot remastering, while model uncertainties were used to theoretically calculate the accuracy of the robotic arm and quantify sources of error as discussed later in Section \ref{sec:theor_acc}.

The Metropolis algorithm generates samples from a probability density function $p(\cdot)$ given a cost function $E(\cdot)$ and a proposal density function $g(\cdot)$. First, an error matrix, $\mathbf{E}$, was defined as the error between all points measured in $N$ by 1) a reference system (${}^N\mathbf{p}_{i,ref} \in \mathbb{R}^{3 \times 1}$), and 2) a robot (${}^N\mathbf{p}_{i,robot} \in \mathbb{R}^{3 \times 1}$),

\begin{equation} \label{eq:error}
\mathbf{E} = {}^N\mathbf{P}_{ref} - {}^N\mathbf{P}_{robot},
\end{equation}

\noindent with a cost function,

\begin{equation} \label{eq:cost_function}
E = \sum_{i=1}^n \mathbf{E}_i \mathbf{E}_i^T,
\end{equation}

\noindent where $\mathbf{E}_i$ is the $i^{th}$ row of $\mathbf{E}$, and ${}^N\mathbf{P}_{ref} = \left[ {}^N\mathbf{p}_{1,ref}, ... ,{}^N\mathbf{p}_{n,ref} \right]^T \in \mathbb{R}^{n \times 3}$ and ${}^N\mathbf{P}_{robot} = \left[ {}^N\mathbf{p}_{1,robot}, ... ,{}^N\mathbf{p}_{n,robot} \right]^T \in \mathbb{R}^{n \times 3}$. All points ${}^N\mathbf{p}_{i,ref}$ were directly measurable, and in this case, $N$ became $M$ or $L$ for data collected from the motion capture system or laser tracker, respectively. On the other hand, ${}^N\mathbf{p}_{i,robot}$ was calculated from

\begin{equation} \label{eq:point_mod_transf}
{}^N\mathbf{p}_{i,robot} = \mathbf{T}\text{*} (\Theta_1) \: {}^N_R{\mathbf{T}} \: {}^R\mathbf{p}_{i,robot} (\Theta_2),
\end{equation}

\noindent where ${}^N_R{\mathbf{T}} \in \mathbb{R}^{4 \times 4}$ was an initial homogeneous transformation matrix, $\mathbf{T}\text{*} (\Theta_1) \in \mathbb{R}^{4 \times 4}$ was a correctional homogeneous transformation matrix, and ${}^R\mathbf{p}_{i,robot} (\Theta_2)$ is the 3D location of a marker as measured by the robot in its current configuration. Specifically, ${}^N_R{\mathbf{T}}$  was calculated using three non-collinear points (sampled from optimization dataset) measured in both reference frames, $N$ and $R$, as in \cite{Marvel2016}. Calculating this registration matrix from only three points yields an erroneous mapping that is exacerbated by the size of the measurement error in both reference frames \cite{van2017strategies}. Since the robot is known to be inaccurate (later quantified in Section \ref{sec:quantifying_accuracy}), and the accuracy of the motion capture system was unknown, corrections to the erroneous ${}^N_R{\mathbf{T}}$ were mandatory to more accurately calculate values for the zero-offsets. Consequently, $\mathbf{T}^\text{*} (\Theta_1)$ was a correctional homogeneous transformation matrix as a function of $\Theta_1 = \lbrace x, y, z, \alpha, \beta, \gamma \rbrace$, where $\lbrace x, y, z \rbrace$ were translational offsets, and $\{\alpha, \beta, \gamma \rbrace$ were Euler ZYX angular offsets. Essentially, $\mathbf{T}\text{*} (\Theta_1)$ made small adjustments to ${}^N_R{\mathbf{T}}$ by simultaneously including the $\Theta_1$ parameters in the optimization process. This addition mitigated the negative effects of an erroneous ${}^N_R{\mathbf{T}}$ via a more complete and parameterized analytical model for ${}^N\mathbf{p}_{i,robot}(\Theta_2)$. Finally, ${}^R\mathbf{p}_{i,robot}(\Theta_2)$ was obtained through the forward kinematics of the robot at the $i^{th}$ joint configuration,

\begin{equation} \label{eq:forward_kinematics}
\begin{array}{ll}
{}^R_{end}{\mathbf{T}}_i (\Theta_2) & = {}^R_1 {\mathbf{T}}_i(\delta\theta_1) \: {}^1_2 {\mathbf{T}}_i(\delta\theta_2) \: ... \: {}^{k-1}_k {\mathbf{T}}_i(\delta\theta_k) \\
{} & = \begin{pmatrix} {}^R_{end}{\mathbf{R}}_i(\Theta_2) & {}^R\mathbf{p}_{i,robot}(\Theta_2) \\ 0 \: 0 \: 0 & 1 \end{pmatrix}
\end{array}
\end{equation}

\noindent for $k$ number of joints and $\Theta_2 = \lbrace \delta\theta_1, \delta\theta_2, ..., \delta\theta_k \rbrace$ were the zero-offset parameters. These zero-offsets were directly added to their respective joint angles along with any additional offset determined by the D-H convention. 

As previously indicated, the Metropolis algorithm required a probability density function which, in this case, is the posterior distribution of the unknown parameters given experimental data. However, deriving a posterior distribution first required a probabilistic error model between ${}^N\mathbf{p}_{i,ref}$ and ${}^N\mathbf{p}_{i,robot}$. Since both the zero-offset and registration correctional offsets have been included in the analytical model for ${}^N\mathbf{p}_{i,robot}$ as in (\ref{eq:point_mod_transf}), only unmodeled effects and true measurement noise remained. The unmodeled effects for a robot (discussed in Section \ref{sec:introduction}) are extremely complex and parameterized models thereof were avoided particularly since their overall contribution to robot accuracy is less than 10 \%. Choosing a simpler approach, the unmodeled effects and measurement noise were summarily lumped into a multi-variate, isotropic Gaussian error model as

\begin{equation} \label{eq:point_error}
{}^N\mathbf{p}_{i,ref} = {}^N\mathbf{p}_{i,robot}+\mathbf{E}_i, \: \mathbf{E}_i \sim N(0,\Sigma).
\end{equation}

\noindent where $\Sigma$ is the covariance matrix

\begin{equation} \label{eq:covariance}
\Sigma = 
\begin{pmatrix}
\sigma^2 & 0 & 0 \\
0 & \sigma^2 & 0 \\
0 & 0 & \sigma^2
\end{pmatrix}.
\end{equation}

\noindent The likelihood of a single data point, ${}^N\mathbf{p}_{i,ref}$, given all parameters is

\begin{equation} \label{eq:likelihood_point}
\begin{array}{ll}
L({}^N\mathbf{p}_{i,ref} | \Theta,\sigma^2) & \propto \frac{1}{\sqrt[]{det(2\pi \Sigma)}}exp\lbrace \frac{-\mathbf{E}_i (\Sigma)^{-1} \mathbf{E}_i^T}{2} \rbrace \\
{} & \propto \frac{1}{\sqrt[]{(2\pi)^3 \sigma^6}}exp\lbrace \frac{-\mathbf{E}_i \mathbf{E}_i^T}{2\sigma^2} \rbrace \\
{} & \propto \sigma^{-3} exp\lbrace \frac{-\mathbf{E}_i \mathbf{E}_i^T}{2\sigma^2} \rbrace,
\end{array}
\end{equation}

\noindent where $\Theta = \lbrace \Theta_1,\Theta_2 \rbrace$. Therefore, the likelihood of all experimental data given all parameters is

\begin{equation} \label{eq:likelihood}
\begin{array}{ll}
L({}^N\mathbf{P}_{ref} | \Theta,\sigma^2) & \propto \displaystyle\prod_{i=1}^{n} L({}^N\mathbf{p}_{i,ref} | \Theta,\sigma^2) \\
{} & \propto \displaystyle\prod_{i=1}^{n} {} \sigma^{-3} exp\lbrace \frac{-\mathbf{E}_i \mathbf{E}_i^T}{2\sigma^2} \rbrace \\
{} & \propto \sigma^{-3n} exp \lbrace \frac{-E}{2\sigma^2} \rbrace.
\end{array}
\end{equation}

The posterior distribution of all parameters given all experimental data is

\begin{equation} \label{eq:bayes}
p(\Theta,\sigma^2 | {}^N\mathbf{P}_{ref}) \propto L({}^N\mathbf{P}_{ref} | \Theta,\sigma^2) \: p(\Theta,\sigma^2),
\end{equation}

\noindent where $p(\Theta,\sigma_1^2)$ is the prior. Combining a flat (constant) prior for $\Theta$ since no information was known regarding these variables \cite{gelman2006prior}, and the typical Jeffrey's prior for an unknown $\sigma$ of a normal distribution \cite{kass1996selection} yielded a joint, non-informative prior, $p(\Theta,\sigma^2) \propto \sigma^{-2}$. Substituting (\ref{eq:likelihood}) and the Jeffrey's prior into (\ref{eq:bayes}) yielded the resulting posterior distribution

\begin{equation} \label{eq:posterior}
p(\Theta,\sigma^2 | {}^N\mathbf{P}_{ref}) \propto  \sigma^{-3n-2} exp \{ -\frac{E}{2\sigma^2} \}.
\end{equation}

\noindent In order to prevent numerical issues when calculating (\ref{eq:posterior}) that arise from typical machine precision and large $n$, the acceptance ratio $\alpha$ for the Metropolis algorithm is calculated as

\begin{equation} \label{eq:acceptance_ratio}
\begin{array}{ll}
\alpha & = \frac{p(\Theta_p,\sigma_p^2 | {}^N\mathbf{P}_{ref})}{p(\Theta,\sigma^2 | {}^N\mathbf{P}_{ref})} \\
{} & = \frac{\sigma_p^{-3n-2} \: exp \{ -\frac{E_p}{2\sigma_p^2} \}}{\sigma^{-3n-2} \: exp \{ -\frac{E}{2\sigma^2} \}} \\
{} & = (\frac{\sigma_p}{\sigma})^{-3n-2} \: exp \{ -\frac{E_p}{2\sigma_p^2} + \frac{E}{2\sigma^2} \}.
\end{array}
\end{equation}

\noindent The posterior distribution (\ref{eq:posterior}) was sampled using symmetric proposal density functions $g(\Theta_p|\Theta) \sim \mathcal{U}(-0.0125,0.0125)+\Theta$ and $g(\sigma_p|\sigma^2) \sim \mathcal{U}(-0.0125,0.0125)+\sigma$ for $2\times10^5$ samples. The initial condition was 1 mm for $\sigma$ (positive value), and 0 mm or deg for the respective parameters in $\Theta$. With the optimization of fourteen parameters, the issue of convergence to local minima exists. This undesirable effect was mitigated by recording positioning data throughout the robot's workspace and configuration space, and with a sufficiently large number of data points as covered in Section \ref{sec:data_collection}. As subsequently discussed, optimization from data of both reference sensors yielded approximately the same parameter values. This mutual convergence provided intrinsic evidence that the global minimum was reached.

\begin{figure}
\centering
\includegraphics[width=1\linewidth]{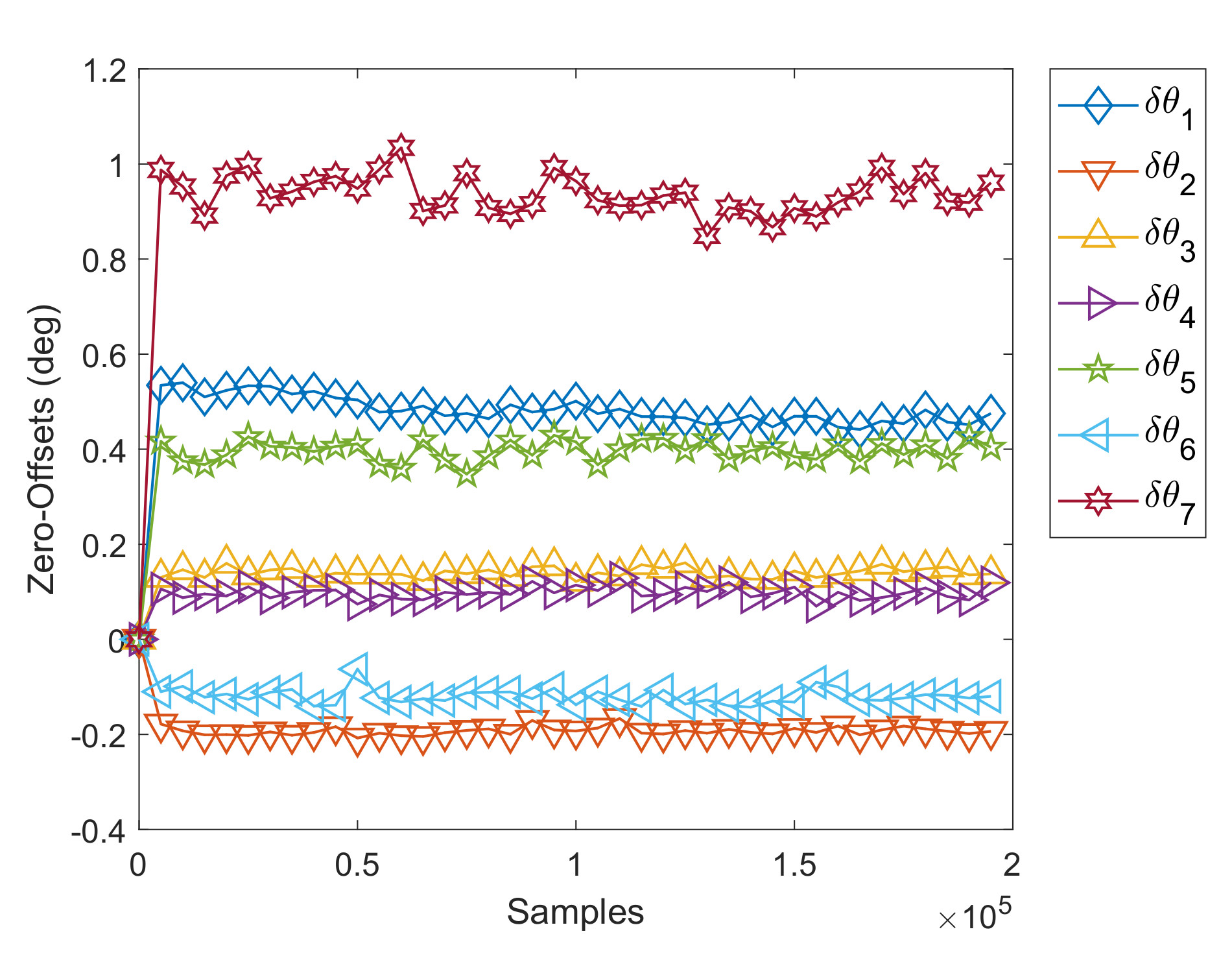}
\caption{Trace plot of zero-offsets of all robot joints during sampling process using motion capture and robot data.}
\label{fig:zero_offset_trace}
\end{figure}

\begin{figure}
\centering
\includegraphics[width=1\linewidth]{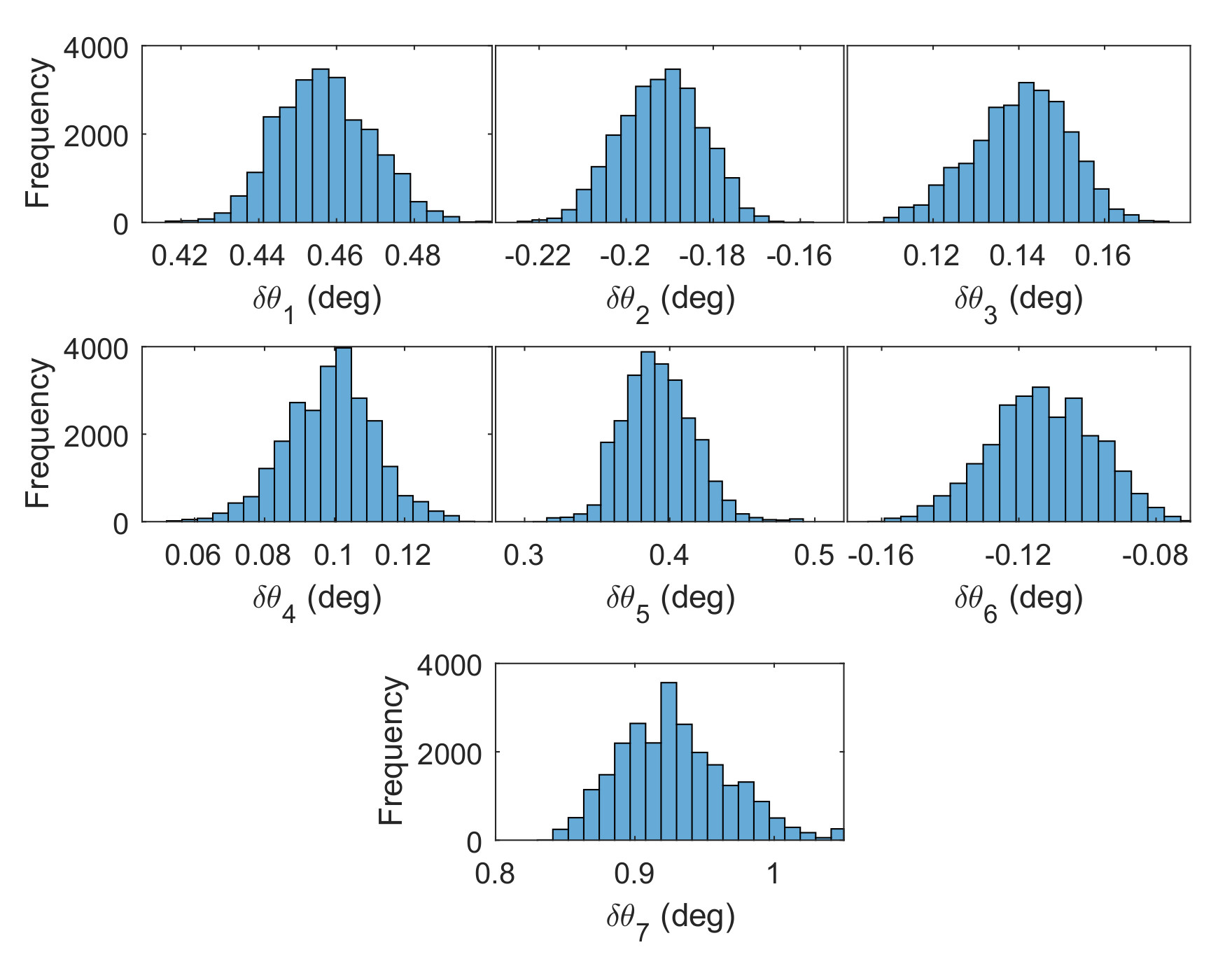}
\caption{Converged zero-offset distributions of all robot joints during the optimization process with motion capture data.}
\label{fig:zero_offset_histograms}
\end{figure}

From $1.75\times10^5$ samples to $2\times10^5$ samples, the joint offsets, $\Theta_2$, surpassed the ``burn-in'' phase and converged as shown in Fig. \ref{fig:zero_offset_trace}. This sampling process captured the most likely estimates (MLE) of the model parameters and their uncertainty. Of particular interest for robot remastering were the zero-offsets and their uncertainty as captured by their corresponding histograms in Fig. \ref{fig:zero_offset_histograms}. By inspection, the zero-offset uncertainties approximately followed a normal distribution, centered at the MLE values. The MLE zero-offsets (calculated as sample means due to unimodal and symmetric distributions in Fig. \ref{fig:zero_offset_histograms}) and their standard deviation are shown in Table \ref{tab:MLE_zero_offsets}. Five MLE offsets were very close in value (within 7 \%) across reference measurement systems and any perceived differences are contained by their uncertainties. True differences (outside parametric uncertainty) were only observed for the first and last zero-offsets: a ramification of the differences in measurement devices and mounting locations of SIR and SMR (the only possible sources that can invoke differences in data generated by the reference devices). Overall, the agreement in optimized parameters provided the first evidence that a significantly lower-cost motion capture measurement system could replace the more conventional laser tracker for remastering robots using the outlined method. Noticeably, the calculated MLE offsets were predominantly very small at fractions of a degree. However, these small changes contribute significantly to the Cartesian accuracy of a robot as revealed in Section \ref{sec:quantifying_accuracy}.

The MLE and standard deviation of the remaining parameters, $\Theta_1$ and $\sigma$, are shown in Table \ref{tab:MLE_other_params}. As expected, the inclusion of the $\Theta_1$ parameters for the calculation of ${}^N \mathbf{p}_{i,robot}$ in (\ref{eq:point_mod_transf}) during optimization significantly adjusted the initial registration matrices with rather large correctional values. In particular, the translational parameters regardless of the reference sensor yielded adjustments of several millimeters. Similar to the zero-offsets, the angular adjustments were sub-degree, but their net-effect had significant effects towards the fidelity of the optimization results and contributed towards the similarity of the converged zero-offsets across reference measurement devices. Finally, $\sigma$ of the isotropic Gaussian error model was slightly higher when calculated from motion capture data than from laser tracker data. This effect is most likely due to the better accuracies exhibited by laser tracker technologies and indicate the magnitude of sources of error that were not captured through the adjustments of either the zero-offsets or the correctional registration matrix. This error estimate, combined with the zero-offsets and their uncertainties, were used to calculate and examine the theoretical positioning accuracy of the robot in Section \ref{sec:theor_acc}.

\begin{table}
\caption{Zero-offsets for all joints obtained with both the laser tracker and motion capture system.}
\centering
\begin{tabular}{|C{1.4cm} |C{1.3cm} | C{1.3cm} | C{1.3cm} | C{1.3cm} |}
\hline
& \multicolumn{2}{c|}{\textbf{Motion Capture}} &  \multicolumn{2}{c|}{\textbf{Laser Tracker}}\\
\cline{1-5}
\bigstrut[t] Zero-Offset & \bigstrut[t] MLE (deg) & \bigstrut[t] \bigstrut[t] Std (deg) & \bigstrut[t] MLE (deg) & \bigstrut[t] Std (deg)\\ 
\hline 
\bigstrut[t] $\delta\theta_1$ & \bigstrut[t] 0.477 & \bigstrut[t] 0.012 &  \bigstrut[t] 0.408 &  \bigstrut[t] 0.009\\
\hline
\bigstrut[t] $\delta\theta_2$ & \bigstrut[t] -0.192 &  \bigstrut[t] 0.010 &  \bigstrut[t] -0.190 & \bigstrut[t] 0.009\\
\hline
\bigstrut[t] $\delta\theta_3$ & \bigstrut[t] 0.139 &  \bigstrut[t] 0.011 &  \bigstrut[t] 0.130 & \bigstrut[t] 0.009\\
\hline
\bigstrut[t] $\delta\theta_4$ & \bigstrut[t] 0.099 &  \bigstrut[t] 0.013 &  \bigstrut[t] 0.105 & \bigstrut[t] 0.013\\
\hline
\bigstrut[t] $\delta\theta_5$ & \bigstrut[t] 0.392 &  \bigstrut[t] 0.024 &  \bigstrut[t] 0.378 & \bigstrut[t] 0.021\\
\hline
\bigstrut[t] $\delta\theta_6$ & \bigstrut[t] -0.114 &  \bigstrut[t] 0.016 &  \bigstrut[t] -0.114 & \bigstrut[t] 0.017\\
\hline
\bigstrut[t] $\delta\theta_7$ & \bigstrut[t] 0.936 &  \bigstrut[t] 0.039 &  \bigstrut[t] 1.262 & \bigstrut[t] 0.026\\
\hline
\end{tabular}
\label{tab:MLE_zero_offsets}
\end{table}

\begin{table}
\caption{Registration correctional parameters and Gaussian error model obtained with the laser tracker and motion capture system data.}
\centering
\begin{tabular}{|C{1.4cm} |C{1.3cm} | C{1.3cm} | C{1.3cm} | C{1.3cm} |}
\hline
& \multicolumn{2}{c|}{\textbf{Motion Capture}} &  \multicolumn{2}{c|}{\textbf{Laser Tracker}}\\
\cline{1-5}
\bigstrut[t] Parameter & \bigstrut[t] MLE & \bigstrut[t] \bigstrut[t] Std & \bigstrut[t] MLE & \bigstrut[t] Std\\ 
\hline 
\bigstrut[t] $x$ (mm) & \bigstrut[t] -2.096 & \bigstrut[t] 0.113 &  \bigstrut[t] 2.240 &  \bigstrut[t] 0.039\\
\hline
\bigstrut[t] $y$ (mm) & \bigstrut[t] 0.502 &  \bigstrut[t] 0.114 &  \bigstrut[t] -0.395 & \bigstrut[t] 0.090\\
\hline
\bigstrut[t] $z$ (mm) & \bigstrut[t] 3.311 &  \bigstrut[t] 0.0748 &  \bigstrut[t] 7.325 & \bigstrut[t] 0.088\\
\hline
\bigstrut[t] $\sigma$ (mm) & \bigstrut[t] 0.890 &  \bigstrut[t] 0.027 &  \bigstrut[t] 0.734 & \bigstrut[t] 0.025\\
\hline
\bigstrut[t] $\alpha$ (deg) & \bigstrut[t] -0.252 &  \bigstrut[t] 0.006 &  \bigstrut[t] -0.918 & \bigstrut[t] 0.010\\
\hline
\bigstrut[t] $\beta$ (deg) & \bigstrut[t] -0.334 &  \bigstrut[t] 0.005 &  \bigstrut[t] 0.099 & \bigstrut[t] 0.005\\
\hline
\bigstrut[t] $\gamma$ (deg) & \bigstrut[t] -0.960 &  \bigstrut[t] 0.017 &  \bigstrut[t] 0.078 & \bigstrut[t] 0.004\\
\hline
\end{tabular}
\label{tab:MLE_other_params}
\end{table}

\section{Quantifying Positional Accuracy}
\label{sec:quantifying_accuracy}
Three different accuracy metrics are proposed to methodically quantify the positioning performance of the arm before and after remastering. These metrics quantified both relative and absolute positioning performance as well as the contribution of error sources thereof. Overall, the proposed Bayesian remastering method yielded a multi-factor improvement in the Cartesian positioning accuracy of the arm when compared with the Vernier remastering method.

\subsection{Relative Cartesian Accuracy}
\label{sec:dist_accuracy_measure}
An easily-obtained relative measure of robot accuracy involved calculating the relative distances between marker locations as measured by the robot and reference system for \textit{all} robot joint configurations of a \textit{single} Cartesian pose. Combining these relative distances across all 36 Cartesian poses yielded 440 distance measurements, quantifying the variability of the robot's Cartesian positioning as a function of its joint configuration solution. Ideally, a robot should not exhibit Cartesian pose variations based on the choice of joint configuration (an assumption used by simulators and kinematic models of robots). The mathematical formulation for this metric is

\begin{equation} \label{eq:dist_equation}
\begin{array}{ll}
d^k_{ij} & = |||{}^N\mathbf{p}^k_{i,ref} - {}^N\mathbf{p}^k_{j,ref} ||_2 - ||{}^R\mathbf{p}^k_{i,robot} - {}^R\mathbf{p}^k_{j,robot}||_2| \\

& \quad i=1,...,n-1, \\
& \quad j=i,...,n \\
& \quad k=1,...,K
\end{array}
\end{equation}

\noindent where ${}^N\mathbf{p}^k_{i,ref}, {}^N\mathbf{p}^k_{j,ref}, {}^R\mathbf{p}^k_{i,robot}, {}^R\mathbf{p}^k_{j,robot} \in C_k$, a cluster of points defined by all joint configurations of a robot for the $k^{th}$ Cartesian pose. Since the distance between two Cartesian points is theoretically invariant of measurement coordinate system, this formulation used both $N$ and $R$. Furthermore, this metric is a lower-bound estimate of robot accuracy as it discards some sources of error that contribute towards absolute positioning accuracy.

The numerical results for this accuracy measure are shown in Table \ref{tab:dist_accuracy_measure}. Both the motion capture system and laser tracker indicated rather large positioning errors with an average error over 3 mm and an upper $97.5^{th}$ percentile over 6 mm with the Vernier remastering method. In contrast, the proposed Bayesian remastering method produced average positioning errors under 0.3 mm with an upper $97.5^{th}$ percentile under 1 mm. The results from the latter method were calculated by updating the kinematic model of the robot (\ref{eq:forward_kinematics}) using the MLE zero-offsets in Table \ref{tab:MLE_zero_offsets} for all experimentally tested robot configurations in the optimization dataset. This process yielded new values for ${}^R\mathbf{p}_{i,robot}$ and ${}^R\mathbf{p}_{j,robot}$ in (\ref{eq:dist_equation}), and new values for $d_{ij}$ were calculated. Estimates indicated a near 10-fold improvement in the average relative Cartesian accuracy with persistent sub-millimeter accuracy even in extreme cases.

Encouragingly, the mean and 95 \% confidence intervals of positioning errors for this distance metric between the motion capture system and laser tracker were very close in agreement. This observation provided further evidence that the motion capture system can, at the very least, accurately perceive distances in space. Unfortunately, since the two types of markers were not mounted in the same location (and therefore not measuring the same point), it is impossible to disentangle the sources of the discrepancies between the two reference sensors. Primary sources influencing measurement discrepancy include the accuracy and precision of the sensors and the different locations of the markers. The latter arose from kinematic errors of the robot propagating differently to the positionally disparate markers at the end-effector. This trait affected all subsequent accuracy estimates.

\begin{table}
\caption{Relative Cartesian accuracy calculated from motion capture and laser tracker data before and theoretically after robot remastering.}
\centering
\begin{tabular}{|C{1.7cm} |C{1.9cm} | C{1.4cm} | C{1.7cm} |}
\hline
\bigstrut[t] Remastering Method & \bigstrut[t] Reference Sensor & \bigstrut[t] Mean (mm) & \bigstrut[t] 95\% Interval (mm)\\ 
\hline 
\multirow{2}{*}{Vernier} & \bigstrut[t] Motion Capture & \bigstrut[t] 3.384 &  \bigstrut[t] [0.788, 6.513]\\
\cline{2-4}
& \bigstrut[t] Laser Tracker &  \bigstrut[t] 3.567 &  \bigstrut[t] [0.673, 6.359]\\
\hline
\multirow{2}{*}{Bayesian} & Motion Capture &  \bigstrut[t] 0.286 &  \bigstrut[t] [0.007, 0.951]\\
\cline{2-4}
& Laser Tracker &  \bigstrut[t] 0.264 &  \bigstrut[t] [0.010, 0.800]\\
\hline
\end{tabular}
\label{tab:dist_accuracy_measure}
\end{table}

\subsection{Post-Registration Cartesian Accuracy}
\label{sec:reg_error_upper}
The primary estimate of the robot's absolute positioning accuracy was calculated as

\begin{equation} \label{eq:post_reg_accuracy}
e = \sum_{i=1}^n ||\mathbf{E}_i||_2
\end{equation}

\noindent across all robot configurations. This metric is sensitive to all sources of error including registration error, joint-offset error, measurement error, and other unmodeled robot factors as discussed in Section \ref{sec:introduction}. Consequently, the actual robot accuracy is less than $e$ as calculated in ($\ref{eq:post_reg_accuracy}$) since a robot's accuracy would only depend on joint-offset error and unmodeled dynamics. As shown in Table \ref{tab:reg_accuracy_measure}, the average absolute positioning error of the robot is over 4 mm with the Vernier remastering method regardless of reference sensor. Lower percentile estimates indicated at least 1 mm of error, while upper percentile estimates indicate over 7 mm of error. As expected, this measure of accuracy is much larger than the values of the preceding relative measure. Regardless, this level of erroneous positioning can negatively impact the quality of operations including offline robot programming, high-tolerance insertions, acquisition of small parts, and placement accuracy of parts. Positively, the proposed Bayesian remastering method yielded at least a three-fold improvement in absolute accuracy across the mean and both the lower and upper percentiles. Again, these estimates were calculated by updating the kinematic model of the robot in (\ref{eq:forward_kinematics}) using the MLE zero-offsets in Table \ref{tab:MLE_zero_offsets} for all experimentally tested robot configurations in the optimization dataset. This process yielded new values for ${}^N\mathbf{p}_{i,robot}$ in (\ref{eq:point_error}) with existing values of ${}^N\mathbf{p}_{i,robot}$, updating estimates for $e$. Although the absolute positioning error of the robot was substantially improved with the Bayesian remastering method, others have reported error values under 1 mm. However, the primary source of this larger error is likely due to the mechanics of this particular robotic arm, and not the remastering method itself as discussed in the following section.

\begin{table}
\caption{Post-registration absolute position error calculated from motion capture and laser tracker data before and theoretically after remastering robot.}
\centering
\begin{tabular}{|C{1.7cm} |C{1.9cm} | C{1.4cm} | C{1.7cm} |}
\hline
\bigstrut[t] Remastering Method & \bigstrut[t] Reference Sensor & \bigstrut[t] Mean (mm) & \bigstrut[t] 95\% Interval (mm)\\ 
\hline 
\multirow{2}{*}{Vernier} & \bigstrut[t] Motion Capture & \bigstrut[t] 4.715 &  \bigstrut[t] [1.508, 8.581]\\
\cline{2-4}
& \bigstrut[t] Laser Tracker &  \bigstrut[t] 4.397 &  \bigstrut[t] [1.517, 7.256]\\
\hline
\multirow{2}{*}{Bayesian} & Motion Capture &  \bigstrut[t] 1.533 &  \bigstrut[t] [0.459, 2.770]\\
\cline{2-4}
& Laser Tracker &  \bigstrut[t] 1.213 &  \bigstrut[t] [0.504, 2.128]\\
\hline
\end{tabular}
\label{tab:reg_accuracy_measure}
\end{table}

\subsection{Theoretical Accuracy and Uncertainty Analysis}
\label{sec:theor_acc}
\begin{figure}
\centering
\includegraphics[width=1\linewidth]{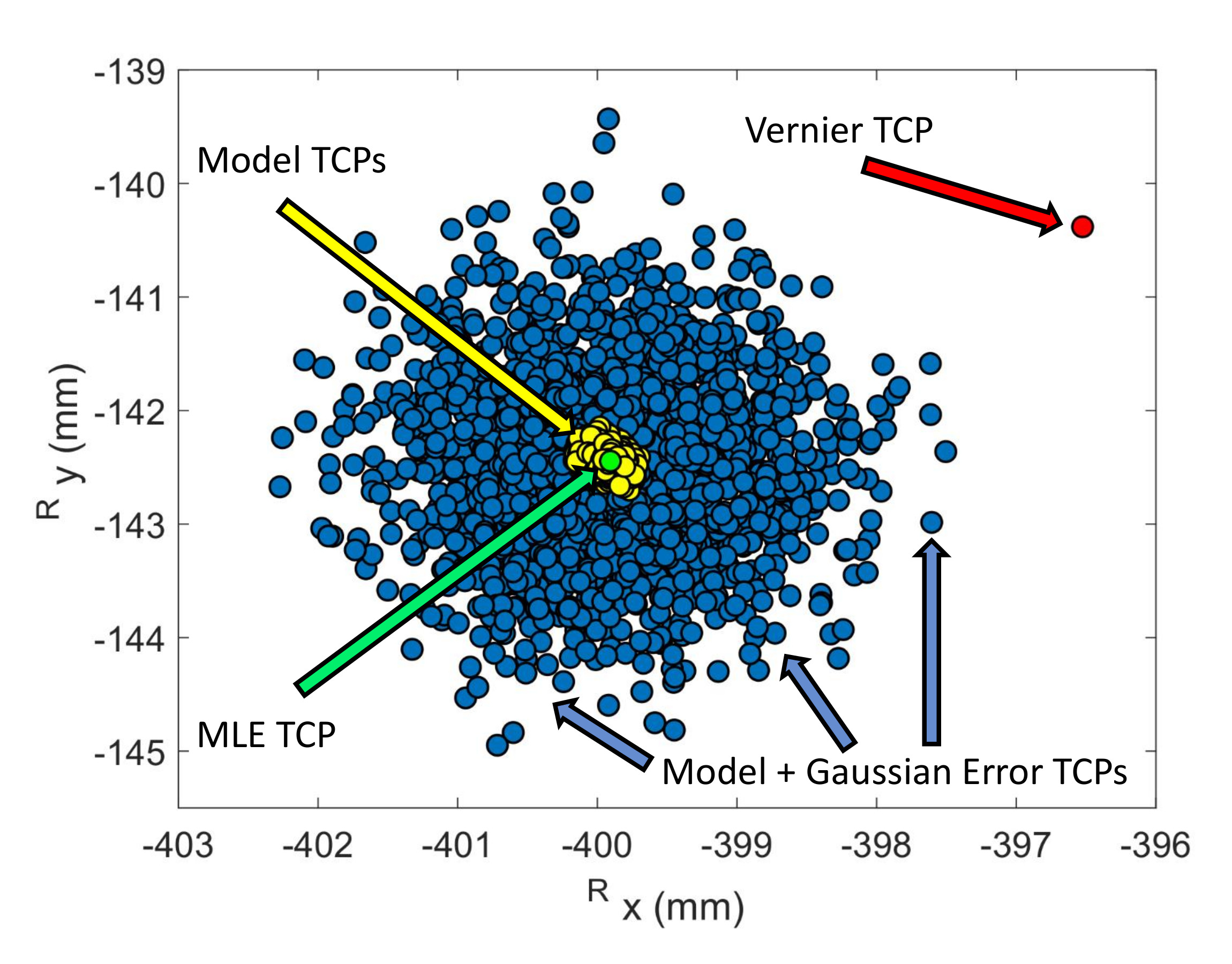}
\caption{Robot TCPs calculated for a single joint configuration using Vernier zero-offsets (red), Bayesian MLE zero-offsets (green), probable Bayesian zero-offsets (yellow), and probable Bayesian zero-offsets with isotropic Gaussian error (blue). Robot TCP is three-dimensional, but only plotted across the X and Y dimensions for visual clarity.}
\label{fig:accuracy}
\end{figure}

A new, theoretical measure of a robot's Cartesian positioning accuracy involved analyzing the uncertainty of the zero-offsets and model error from the optimization process. As previously discussed, the vast majority of error associated with the positional accuracy of a robotic arm stems directly from erroneous zero-offsets of the robot's joints. Therefore, analyzing error effects of the zero-offsets via their uncertainty conveniently paved a path for theoretically quantifying an arm's positional accuracy using (\ref{eq:forward_kinematics}). In practice, the MLE zero-offsets obtained were applied to the physical robot arm, but adopting these values did not guarantee that the correct zero-offsets were actually chosen. In fact, the correct zero-offsets could, in theory, be any of the values generated by the Metropolis sampling process as captured in Fig. \ref{fig:zero_offset_histograms}. Therefore, to estimate the positional accuracy (theoretically) of the robot pre- and post-optimization, end-effector positions (${}^R \mathbf{p}_{i,robot}$, or more simply, TCP) are generated for the arm in (\ref{eq:forward_kinematics}) for the joint configurations visited in the optimization dataset using the 1) Vernier zero-offsets (Vernier TCP), 2) Bayesian MLE zero-offsets (MLE TCP), and 3) 2000 randomly selected, Bayesian zero-offsets per joint (Model TCPs) with and without the isotropic Gaussian error model. The identified isotropic Gaussian error model in (\ref{eq:point_error}) was sampled and directly added to ${}^R \mathbf{p}_{i,robot}$ due to its relevance in capturing positioning accuracy. Its direct addition is possible since the error model is isotropic, and therefore, its effects are not changed through expression in different Cartesian coordinate systems. Fig. \ref{fig:accuracy} illustrates the TCP positions for these cases for a single arm configuration.

For each arm configuration, the Euclidean distance is calculated between the TCPs generated using the MLE zero-offsets and the 2000 TCPs generated from randomly selected, probable zero-offsets with isotropic error. These distances indicated the error in positioning the TCP of the robotic arm assuming 1) MLE zero-offsets were applied to the robot, 2) the MLE zero-offsets were incorrect, and 3) the correct zero-offsets were among those values as shown in Fig. \ref{fig:zero_offset_histograms}. Furthermore, the Euclidean distances were also calculated between the TCP generated with Vernier TCP offsets and the 2000 TCPs generated from randomly-selected, probably zero-offsets with isotropic error. These distances indicated the error in TCP positioning assuming 1) only Vernier zero-offsets were used, and 2) the correct zero-offsets were among those generated by the Metropolis process.

\begin{table}
\caption{Theoretical absolute position error calculated from the uncertainty of zero-offsets, unmodeled dynamics, and noise at optimization joint configurations.}
\centering
\begin{tabular}{|C{1.7cm} |C{1.9cm} | C{1.4cm} | C{1.8cm} |}
\hline
\bigstrut[t] Remastered & \bigstrut[t] Reference Sensor & \bigstrut[t] Mean (mm) & \bigstrut[t] 95\% Interval (mm)\\ 
\hline 
\multirow{2}{*}{Vernier} & \bigstrut[t] Motion Capture & \bigstrut[t] 4.730 &  \bigstrut[t] [1.353, 8.470]\\
\cline{2-4}
& \bigstrut[t] Laser Tracker &  \bigstrut[t] 4.410 &  \bigstrut[t] [1.673, 7.737]\\
\hline
\multirow{2}{*}{Bayesian} & Motion Capture &  \bigstrut[t] 1.433 &  \bigstrut[t] [0.417, 2.750]\\
\cline{2-4}
& Laser Tracker &  \bigstrut[t] 1.181 &  \bigstrut[t] [0.344, 2.265]\\
\hline
\end{tabular}
\label{tab:theor_accuracy_measure_all_opt_configs}
\end{table}

The mean and 95 \% confidence interval of this theoretical positioning error is shown in Table \ref{tab:theor_accuracy_measure_all_opt_configs}. This metric estimated that the average positioning error of the robot with Vernier remastering can be over 4 mm regardless of reference sensor. Furthermore, the lower percentile guarantees at least 1 mm of positioning error, while the upper percentile of positioning error can be over 7 mm. These error estimates are quite close to those reported in Table \ref{tab:reg_accuracy_measure}, indicating that the theoretical model corroborated the experimental data. Again, the magnitudes of these errors are quite significant in the context of many applications. Fortunately, after the zero-offsets obtained from Bayesian remastering were theoretically applied, this method estimated average positioning errors within 1.5 mm and with upper percentile estimates under 3 mm, the same as those reported in Table \ref{tab:reg_accuracy_measure}.

With a relatively small number of Cartesian positions in the optimization dataset, a more accurate estimate of the robot's true positioning accuracy could be obtained with a large number of configurations. The validated robot model was sampled as before, but at 1000 joint configurations instead of just those configurations visited in the optimization dataset (173 configurations). Calculating the theoretical positioning error from this dataset yielded positioning accuracies as reported in Table \ref{tab:theor_accuracy_measure_all}. In this case, the average positioning error is well over 5 mm with the Vernier remastering method with upper percentiles over 10 mm. The likely cause for this error inflation stems from sampling arm configurations well outside the dexterous workspace (the volume used in experiments) to the full workspace. In essence, inaccuracies in the joint offsets, particularly angular errors, propagate to large Cartesian errors when the arm is in the outer reaches of its workspace. However, the Bayesian remastering method yielded average positioning errors under 1.5 mm and upper percentile errors under 3 mm - similar errors to those in Tables \ref{tab:reg_accuracy_measure} and \ref{tab:theor_accuracy_measure_all_opt_configs}. Consequently, this robot and error model indicated that the absolute positioning accuracy of the robot arm was improved at least four-fold within its \textit{full} workspace.

Finally, the positioning error due to uncertainty in the zero-offsets alone was isolated to examine the contribution of this source of error. Model TCPs were calculated as previously described and were plotted in yellow in Fig. \ref{fig:accuracy}. Model TCPs contain the positioning error due to the uncertainty in zero-offsets alone and their effect on Cartesian accuracy is shown in Table \ref{tab:theor_accuracy_measure}. In comparison to those results in Table \ref{tab:theor_accuracy_measure_all}, the mean, lower, and upper percentile errors are only slightly lower after the robot was remastered with the Vernier method. Specifically, the Vernier method produced robot theoretical absolute position errors of: 1)  6.067 mm (from motion capture) and 5.575 mm (from laser tracker) from the error in joint-offsets alone; and 2) 6.331 mm (from motion capture) and 5.676 mm (from laser tracker) from the error in joint-offsets, unmodeled dynamics, and noise. Therefore, the error in zero-offsets alone accounted for over 95 \% of Cartesian positioning error (literature indicated over 90 \%) with Vernier remastering. This corroborated the widely-reported notion that zero-offsets typically dominate positioning error prior to their optimization. In contrast, the Bayesian remastering method produced robot theoretical absolute position errors of: 1) 0.223 mm (from motion capture) and 0.183 mm (from laser tracker) from the error in zero-offsets alone; and 2) 1.439 mm (from motion capture) and 1.187 mm (from laser tracker) from the error in joint-offsets, unmodeled dynamics, and noise. Therefore, the errors in zero offsets alone accounted for only 15 \% of the final Cartesian error, an effect that is visually discernible from Fig. \ref{fig:accuracy}. Unfortunately, the majority of positioning error after Bayesian remastering stemmed from unmodeled dynamics, errors in other D-H parameters, and noise - effects that cannot be compensated for solely with zero-offsets.

\begin{table}
\caption{Theoretical absolute position error calculated from the uncertainty of zero-offsets, unmodeled dynamics, and noise at 1000 randomly generated joint configurations.}
\centering
\begin{tabular}{|C{1.7cm} |C{1.9cm} | C{1.4cm} | C{1.8cm} |}
\hline
\bigstrut[t] Remastered & \bigstrut[t] Reference Sensor & \bigstrut[t] Mean (mm) & \bigstrut[t] 95\% Interval (mm)\\ 
\hline 
\multirow{2}{*}{Verier} & \bigstrut[t] Motion Capture & \bigstrut[t] 6.331 &  \bigstrut[t] [2.041, 10.908]\\
\cline{2-4}
& \bigstrut[t] Laser Tracker &  \bigstrut[t] 5.676 &  \bigstrut[t] [1.795, 10.191]\\
\hline
\multirow{2}{*}{Bayesian} & Motion Capture &  \bigstrut[t] 1.439 &  \bigstrut[t] [0.420, 2.758]\\
\cline{2-4}
& Laser Tracker &  \bigstrut[t] 1.187 &  \bigstrut[t] [0.345, 2.276]\\
\hline
\end{tabular}
\label{tab:theor_accuracy_measure_all}
\end{table}

\begin{table}
\caption{Theoretical absolute position error calculated from the uncertainty of zero-offsets at 1000 randomly generated joint configurations.}
\centering
\begin{tabular}{|C{1.7cm} |C{1.9cm} | C{1.4cm} | C{1.8cm} |}
\hline
\bigstrut[t] Remastered & \bigstrut[t] Reference Sensor & \bigstrut[t] Mean (mm) & \bigstrut[t] 95\% Interval (mm)\\ 
\hline 
\multirow{2}{*}{Vernier} & \bigstrut[t] Motion Capture & \bigstrut[t] 6.067 &  \bigstrut[t] [1.918, 10.385]\\
\cline{2-4}
& \bigstrut[t] Laser Tracker &  \bigstrut[t] 5.575 &  \bigstrut[t] [1.865, 9.681]\\
\hline
\multirow{2}{*}{Bayesian} & Motion Capture &  \bigstrut[t] 0.223 &  \bigstrut[t] [0.052, 0.512]\\
\cline{2-4}
& Laser Tracker &  \bigstrut[t] 0.183 &  \bigstrut[t] [0.042, 0.436]\\
\hline
\end{tabular}
\label{tab:theor_accuracy_measure}
\end{table}

\section{Validation}
\label{sec:validation}
Thus far, measurements of robot accuracy have only been conducted on the robot configurations in the optimization dataset and theoretical configurations. To assess the degree to which these previous estimates of robot accuracy remain true, the metrics were re-calculated on a validation dataset. To begin, the MLE zero-offsets that were calculated from motion capture system data were directly applied to the robot using its supplied interface for remastering. Unfortunately, MLE zero-offsets from both reference sensors could not be separately applied and assessed due to a design limitation of the robot's remastering interface that prevented the ability to completely rollback to a previously remastered state. Since this research was concerned with providing a low-cost solution to robot remastering, only the MLE zero-offsets from the motion capture system were used.

As indicated in Section \ref{sec:data_collection}, a different set of 32 valid Cartesian poses (producing 156 unique robot joint configurations) was created to reflect a validation dataset. The distance and post-registration accuracy measures were re-calculated using the newly acquired experimental data, and the results are shown in Table \ref{tab:validation_measures}. In reality, performance obtained on a training dataset will almost always reveal better results than on a validation dataset. As expected, this was also the case across all relevant measures of robot accuracy. Relative positioning error was around 0.5 mm with upper limits slightly over 1.1 mm. Although inflated in comparison to the Bayesian remastered errors in Table \ref{tab:dist_accuracy_measure}, these validation errors still indicate that the relative positioning accuracy was primarily sub-millimeter which is consistent with the findings in Section \ref{sec:dist_accuracy_measure}. More modest inflations were seen in the average absolute positioning error: absolute errors from the validation set were on average close to 1.5 mm and under 3 mm in extreme cases, which is consistent with the absolute errors reported in Tables \ref{tab:reg_accuracy_measure} - \ref{tab:theor_accuracy_measure_all}. With such subtle differences among accuracy measures calculated between the optimization and validation datasets, the improvement in Cartesian robot positioning was conclusive, and the sizes of datasets were sufficiently large to produce consistent results.

\begin{table}
\caption{Relative and post-registration absolute error calculated from the validation dataset.}
\centering
\begin{tabular}{|C{1.9cm} |C{1.9cm} | C{1.4cm} | C{1.8cm} |}
\hline
\bigstrut[t] Accuracy Metric & \bigstrut[t] Reference Sensor & \bigstrut[t] Mean (mm) & \bigstrut[t] 95\% Interval (mm)\\ 
\hline 
\multirow{2}{*}{Relative} & \bigstrut[t] Motion Capture & \bigstrut[t] 0.558 &  \bigstrut[t] [0.128, 1.21]\\
\cline{2-4}
& \bigstrut[t] Laser Tracker &  \bigstrut[t] 0.492 &  \bigstrut[t] [0.113, 1.133]\\
\hline
\multirow{2}{2cm}{\centering Post-Registration Absolute} & Motion Capture &  \bigstrut[t] 1.838 &  \bigstrut[t] [0.794, 2.952]\\
\cline{2-4}
& Laser Tracker &  \bigstrut[t] 1.509 &  \bigstrut[t] [0.696, 2.556]\\
\hline
\end{tabular}
\label{tab:validation_measures}
\end{table}

\section{Discussion}
\label{sec:discussion}
This research prioritized the development and performance measurement of an efficient method for remastering robotic manipulators. By design, the new remastering method combined a commercially available, low-cost motion capture system and Bayesian inference to quickly remaster a robot \textit{in situ} without the need of high-accuracy marker plates fixed to the base of the robot. The data collection process outlined a strategy for effectively sampling and measuring the location of markers fixed to the end of the arm within its dexterous workspace. The process mitigated the likelihood of marker visual occlusion with respect to their measurement systems while increasing the likelihood that candidate Cartesian poses were kinematically feasible. The optimization process leveraged Bayesian probability theory, robot kinematics, sensor-robot registration, and an isotropic Gaussian error model to generate candidate zero-offsets that improve the robot's Cartesian accuracy. This effort principally focused on refining the zero-offsets as the sole mechanism for improving robot positioning since many existing robot systems harbor native interfaces for updating the values of these parameters. In turn, the zero-offsets can be seamlessly applied to existing robot systems once they are calculated.

Three different metrics were leveraged to capture the positioning performance of the robot after remastering with the supplied Vernier method and the proposed Bayesian method. The first metric captured the relative positioning variability of a robot as a function of all its joint configurations for a single Cartesian pose. As a relative measure, this metric is a lower-bound to the absolute positioning accuracy of a robot. However, it was simple to calculate and can immediately signify the magnitude of positioning inaccuracy. Overall, the Bayesian method improved the relative Cartesian positioning accuracy of a KUKA robot arm at least six-fold with sub-millimeter accuracy over the supplied Vernier-based remastering method. The second metric quantified the post-registration absolute positioning error of the arm with experimental data. This metric revealed that the Bayesian method improved the absolute Cartesian positioning accuracy nearly three-fold within its dexterous workspace. Finally, the third metric theoretically quantified the absolute positioning error of the arm. Calculations indicated that the Bayesian remastering method improved the absolute positioning error nearly four-fold within its entire workspace when compared to the Vernier method. This theoretical assessment was calculated with experimentally-validated kinematic and error models.

Unfortunately, the average absolute positioning accuracy of the robot could not be improved to a sub-millimeter level of performance, an often reported result in the literature. This level of improvement is not unprecedented as others have shown sub-\textit{centimeter} accuracy after remastering with a laser tracker and a proven mathematical approach for remastering robots \cite{Choi2016}. Ultimately, the fidelity of robot remastering depends not only on the method and reference sensors, but also on the traits of the robots. Indeed, analysis of the uncertainty of a robot kinematic model and Gaussian error indicated that only 15 \% of the remaining error (average near 1.5 mm) could be explained by the uncertainty in zero-offsets. Consequently, very little improvement in positioning accuracy could be obtained with even more refined zero-offsets. In effect, one could conclude that the majority of remaining positioning error is not related to zero-offsets, but rather other mechanical sources including errors in the other kinematic D-H parameters (e.g., link lengths) and structural deflection in the links and joints (often reported as the second most dominant factor in robot positioning accuracy). Unfortunately, many existing robots do not provide intrinsic methods or access for compensating for these effects likely due to their increased complexity and relatively insignificant contribution to the overall positioning accuracy when compared to joint zero-offsets. Regardless,  reported sub-millimeter accuracies were obtained with industrial robots and either expensive laser trackers or non-commercial reference sensors, all of which were influential sources towards remastering fidelity. Conclusively, this new method evinces a competitive price-performance remastering approach in terms of instrument cost, remastering time, pure zero-offset optimization (widely accessible for existing robots), and convenience with \textit{in situ} data collection.

Finally, this research was also concerned with evaluating the consistency of a low-cost motion capture system against a laser tracker, the gold standard towards robot remastering. The optimization process revealed that the MLE joint-offsets and their uncertainties were very similar when calculated from data captured by the motion capture system and laser tracker, separately. Moreover, the various calculations of robot accuracy were relatively consistent between both reference sensors, albeit with slight differences that can also be attributed to differences in marker placement. This research confirmed that a low-cost motion capture system is a viable reference sensor for remastering robots.

\section*{Disclaimer}
\label{sec:disclaimer}
Certain commercial equipment, instruments, or materials are identified in this paper to foster understanding. Such identification does not imply recommendation or endorsement by the National Institute of Standards and Technology, nor does it imply that the materials or equipment identified are necessarily the best available for the purpose.

\bibliographystyle{IEEEtran}
\bibliography{references.bib}

\end{document}